# STOCHASTIC OPTIMIZATION APPROACHES FOR SOLVING SUDOKU

Meir Perez and Tshilidzi Marwala

*School of Electrical & Information Engineering, University of the Witwatersrand, Private Bag 3, 2050, Johannesburg, South Africa*

Phone: +27117177217

Fax: +27114031929

E-mail: t.marwala@ee.wits.ac.za

Abstract

In this paper the Sudoku problem is solved using stochastic search techniques and these are: Cultural Genetic Algorithm (CGA), Repulsive Particle Swarm Optimization (RPSO), Quantum Simulated Annealing (QSA) and the Hybrid method that combines Genetic Algorithm with Simulated Annealing (HGASA). The results obtained show that the CGA, QSA and HGASA are able to solve the Sudoku puzzle with CGA finding a solution in 28 seconds, while QSA finding a solution in 65 seconds and HGASA in 1.447 seconds. This is mainly because HGASA combines the parallel searching of GA with the flexibility of SA. The RPSO was found to be unable to solve the puzzle.

## 1. Introduction

Sudoku is a popular logic-based combinatorial puzzle game [1-6]. A Sudoku puzzle consists of 81 cells, contained in a 9x9 grid. Each cell can contain a single integer ranging between one and nine. The grid is further split up into nine 3x3 sub-grids. The purpose of Sudoku is to fill up the entire 9x9 grid such that the following constraints are met [1]:

- Each row of cells is only allowed to contain the integers one through to nine exactly once.
- Each column of cells is only allowed to contain the integers one through to nine exactly once.
- Each 3x3 sub-grid is also only allowed to contain the integers one through to nine exactly once.

A number of cells in the grid are pre-defined by the puzzle setter, resulting in the Sudoku puzzle having a single, unique solution. Figure 1 depicts a typical Sudoku puzzle and these puzzles are set at different difficulties. The puzzle in Figure 1 is an easy puzzle since it can be solved logically and does not require guessing and its solution is depicted in Figure 2.

| 1 |   |   |   |   |   |   |   | 2 |
|---|---|---|---|---|---|---|---|---|
|   |   | 8 |   |   | 9 |   | 3 | 7 |
| 7 |   |   | 5 | 3 |   |   | 8 |   |
|   | 8 |   |   | 7 | 3 |   | 5 | 4 |
|   |   | 6 | 4 |   | 2 | 7 |   |   |
| 9 | 7 |   | 8 | 5 |   |   | 1 |   |
|   | 1 |   |   | 8 | 7 |   |   | 9 |
| 3 | 4 |   | 6 |   |   | 8 |   |   |
| 8 |   |   |   |   |   |   |   | 1 |

Figure 1: An easy Sudoku Puzzle

| 1 | 5 | 3 | 7 | 6 | 8 | 9 | 4 | 2 |
|---|---|---|---|---|---|---|---|---|
| 4 | 6 | 8 | 1 | 2 | 9 | 5 | 3 | 7 |
| 7 | 2 | 9 | 5 | 3 | 4 | 1 | 8 | 6 |
| 2 | 8 | 1 | 9 | 7 | 3 | 6 | 5 | 4 |
| 5 | 3 | 6 | 4 | 1 | 2 | 7 | 9 | 8 |
| 9 | 7 | 4 | 8 | 5 | 6 | 2 | 1 | 3 |
| 6 | 1 | 5 | 3 | 8 | 7 | 4 | 2 | 9 |
| 3 | 4 | 2 | 6 | 9 | 1 | 8 | 7 | 5 |
| 8 | 9 | 7 | 2 | 4 | 5 | 3 | 6 | 1 |

Figure 2: The solution to the Sudoku Puzzle in Figure 1

Various algorithms have been implemented to solve the Sudoku problem [1]. For the Sudoku puzzle in Figure 1, a logic-based algorithm, mimicking the way a human would solve the puzzle, is adequate to attain a solution. Harder puzzles, where guessing is required, can be solved using backtracking algorithms [1]. The problem with backtracking is that the efficiency of the algorithm is dependent on the number of guesses required to solve the puzzle [7]. Hence, harder puzzles will take longer to solve. A solution to this problem could lie in using stochastic optimization techniques.

Some work has been done on solving Sudoku using stochastic optimization techniques [6-8]. The primary motivation behind using these techniques is that difficult puzzles can be solved as efficiently as simple puzzles. This is due to the fact that the solution space is searched stochastically until a suitable solution is found [6]. Hence, the puzzle does not have to be logically solvable or easy for a solution to be reached efficiently.

Furthermore, stochastic optimization techniques are used to find the global optimum of a problem which contains many local optima [8]. Due to the constraint nature of Sudoku, it is very likely to find a solution which satisfies some of the constraints but not all of them, hence finding a local optimum. Due to the stochastic nature of these techniques, the solution space is still searched even though a local optimum has been found, allowing for the global optimum to be detected.

This paper explores the implementation of four different stochastic optimization techniques as applied to the Sudoku problem: Cultural Genetic Algorithm, Repulsive Particle Swarm Optimization, Quantum Simulated Annealing and Hybrid Genetic Algorithm and Simulated Annealing. Each Technique is discussed, implemented and tested on the puzzle in Figure 1. Results are compared and critically evaluated.

All four techniques share some common features. Each technique requires an initialization process, where the solution space is defined, as well as a fitness function, which ensures that the objectives and the constraints are being adhered to.

## 2. Solution Space Representation

The puzzle in Figure 1 contains 47 empty cells, corresponding to the solution space. There are three ways the solution space can be represented. The first way is to treat each one of the 47 cells as a separate variable, individual or particle, with each individual/particle requiring its own population. Therefore, the solution space consists of 47 separate population groups. The problem with this approach is that each individual/particle can only be operated upon separately. This prevents the possibility of interaction between individuals/particles and also is more computationally demanding.

The second approach is to treat a combination of 47 integers ranging between 1 and 9 (corresponding to the empty cells in figure 1) as one individual/particle. Therefore, the solution space consists of one population with each individual/particle having 47 dimensions or genes. This approach allows for greater interaction amongst individuals/particles since algorithmic operations are carried out between possible solutions. This approach is also computationally less demanding since there is only one population group.

The third approach is to represent an individual as a puzzle with all its cells filled while ensuring that one of the constraints mentioned above is always met. Thus, when a population or state is initialized, it is ensured that each 3x3 grid in each puzzle contains the numbers 1 to 9 exactly once [2]. Furthermore, any operation carried out on an individual must ensure that this constraint is not violated. This approach is also computationally less demanding (when compared to the first approach) since an individual is still represented as a complete puzzle (as opposed to one cell).

Therefore, only the second and third approaches are considered here: Cultural Genetic Algorithm is implemented using both approaches, Repulsive Particle Swarm Optimization uses the second approach, Quantum Simulated Annealing uses the third approach and Hybrid Genetic Algorithm and Simulated Annealing also uses the third approach.

## 3. Fitness Function

A number of possibilities exist with regards to implementing a suitable fitness function. From an arithmetic perspective, the sum of each column, row and grid must equal 45 and its product must equal 362880. One possible fitness function implements these arithmetic operations to ensure that the constraints are met.

The problem with using arithmetic operation to uphold constraints is that the non-repetition of an integer in the same column, row or grid is not guaranteed: a row containing nine entries of 5 still adds up to 45. This could cause the algorithm to converge to a local minimum and not meet all the constraints. Therefore a different approach is taken.

The fitness function implemented here involves determining whether an integer is repeated or is not present in a particular row, column or grid. A fitness value is assigned to a possible solution based on the number of repeated or non-present integers. The more repeated or non-present integers there are in a solution's rows and columns, the higher the fitness value assigned to that solution.

If the third approach to solution space representation mentioned in section 2 is used, then only repetitions in rows and columns are considered. If the second approach is used then repetitions in the grids also contribute to the fitness value.

## 4. Cultural Genetic Algorithm

Cultural Genetic Algorithm (CGA) is an evolutionary optimization technique where individuals are influenced both genetically as well as culturally [9]. A background to both regular Genetic Algorithm (GA) as well as CGA is presented. The application of CGA to the Sudoku problem is then discussed and two alternative approaches are considered.

### 4.1. Genetic Algorithm Background

GA optimization is a population based optimization technique inspired by biological genetics and the Darwinian theory of evolution (survival of the fittest and natural selection). A good introduction to GA is given in [10]. A population (set of numerical data) is chosen for natural selection. The population comprises a number of potential solutions to a specific problem. Each possible solution is referred to as an individual. An individual consists of a combination of genes. The optimal combination of genes could lie dormant amongst the population and could come from a combination of individuals. An individual with a genetic combination close to the optimal is described as being fit.

A new generation of individuals are created by mating two individuals from the current population. The fitness function is used to determine how close an individual is to the optimal solution. The selection function ensures that genetic information from the fittest individuals is passed down to the next generation, generating a fitter population. Eventually the population will converge on the optimal solution or get as close to it as possible. Implementation of a GA is carried out in four steps: Initialization, selection, reproduction and termination.

*Initialization:* entails encoding the chromosomes into a format suitable for natural selection. There are several types of encoding modalities, each with their advantages and disadvantages. Each individual of a population can be represented as a binary number. Since a binary number consists of ones and zeros (base 2), more digits are required to define an individual than if a decimal number was used (base 10). This lends itself to greater diversity in chromosome representation and hence greater variance in subsequence generations [10]. The problem with binary encoding is that most populations are not naturally represented in binary form due to the length of binary numbers; they are computationally expensive.

Another form of encoding is floating point encoding. Each individual is represented as a floating point number or a combination of floating point numbers. Floating point encoding is far more efficient than binary encoding [10]. Value encoding is similar but allows for characters and commands to represent an individual.

*Selection:* Selection of individuals for mating involves using a fitness function. A fitness function is used to determine how close an individual is to the optimal solution. The fitness function is the only part of the GA which has knowledge of the problem [10]. The fitness function for the Sudoku problem is discussed in Section 3.

After defining the fitness of each individual, it is necessary to select individuals for mating. There are various methods used. Two methods are discussed here. The Roulette technique involves first summing the fitness's of all the individuals of a population and then selecting a random number between zero and the summed result. The fitness's are then summed again until the random number is reached or just exceeded. The last individual to be summed is selected.

Another selection technique is the tournament method. The tournament method involves selecting a random number of individuals from the population and the fittest individual is selected. The larger the number of individuals selected, the better the chance of selecting the fittest individual.

Selection ensures that the fittest individuals are more likely to be chosen for mating but also allows for less fit individuals to be chosen. A selection function which only mates the fittest individuals is termed elitist and may result in the algorithm converging to a local minimum.

*Reproduction*: comprises two different genetic operations: crossover and mutation.

Crossover is the process by which two individuals share their genes, giving rise to a new individual. Crossover ensures that genes of fit individuals are mixed in an attempt to create a fitter new generation. There are various types of crossover depending on the encoding type, two of which are mentioned here: simple and arithmetic crossover.

Simple crossover is carried out on a binary encoded population. This involves choosing a particular point and all genes up until that point will come from the one parent while the rest comes from the other. For example, one parent has the following binary configuration: 11010100. The other has the following 00101111. If the $5^{th}$ point is chosen then the resultant child will have the following configuration: 11010111. It is also possible to choose multiple points, which signify where crossover occurs.

In arithmetic crossover a new generation is created from adding a percentage of one individual to another. For example an individual has the value 9.3 and another 10.7. If we select 30% from the one and 70% from the other the child will be 10.2.

Over the course of reproduction, a child's chromosome will go through mutation. Mutation is when the gene sequence of a chromosome is altered slightly, either by changing a gene or by changing the sequence. This is done to ensure that the population converge to a global minimum as opposed to a local minimum.

*Termination*: determines the criteria for the algorithm to stop. This can be once the optimal solution is reached but could be computationally expensive. Otherwise the GA can terminate once a certain number of generations have been reached, if the optimal solution has not been reached or once no better solution can be achieved.

## 4.2. Cultural Genetic Algorithm Background

Cultural Genetic Algorithm (CGA) is a variant of GA which includes a belief space [9]. There are various categories of belief space [9]:

- Normative belief: where there is a particular range of values to which an individual is bound.
- Domain specific belief: where the information about the domain of the problem is available.
- Temporal belief: where information about important events in search space is available.
- Spatial belief: where the topographical information of the search space is available.

In addition to the belief space, an influence function is also required in order to implement a CGA [9]. An influence function forms the interface between the population and the belief space. It alters individuals in the population to conform to the belief space.

CGA is a suitable candidate for solving Sudoku since it prevents the algorithm from producing an individual which violates the belief space. This reduces the number of possible individuals the GA needs to generate until the optimum is found. Two different approaches are considered, depending on the way the solution space is represented.

## 4.3. CGA Applied to the Sudoku Problem: First Approach

This approach uses the second solution space representation scheme mentioned in Section 2: each individual consists of

47 genes, with each gene corresponding to a non-fixed cell in the puzzle. A population of 50 individuals is randomly initialized, ensuring that each individual contains genes which conform to the belief space.

The belief space consists of three main categories:

- Normative belief: each individual can only contain genes ranging between 1 and 9.
- Domain specific belief: each individual can only contain genes which are integers.
- Spatial belief: each individual can only contain genes that do not result in the repetition of a fixed cell values within the same row, column or grid, as defined in Figure 1.

The third belief criterion contains knowledge of the topography of the search space, in other words the fixed cell values. An influence function is used to ensure that the belief space is adhered to. The influence function ensures that only random numbers between 1 and 9 are used to initialize the puzzle. It also implements a rounding function to ensure that the values are all integers. It also performs a check, to make sure that the random numbers generated are not repetitions of one of the fixed numbers in the same column, row and grid.

Once the problem is initialized, the fitness of each individual in the population is determined using the fitness function described in Section 3 above. A sub-population of 25 individuals is selected for reproduction. A tournament selection is carried on the individuals of the sub-population, to determine which individuals mate. Tournament selection is carried out since it is efficient to code, works on parallel architectures and allows the selection pressure to be easily adjusted [10].

During reproduction both crossover as well as mutation is carried out. Single point, simple crossover is implemented: a number between 1 and 47 is randomly generated from a Gaussian distribution [10]. This number corresponds to the point of crossover. All genes before this point come from the one parent while the other parent contributes the rest. A new individual, whose genetic makeup is a combination of both parents', is thus reproduced.

The new individual also undergoes mutation: three random genes are selected for mutation. These genes are allocated new random values, which still conform to the belief space. The new individuals replace the individuals in the population which have the lowest fitness values. The CGA is run until an individual with a fitness of zero is found, indicating that the solution to the puzzle has been reached.

### 4.4. CGA Applied to the Sudoku Problem: Second Approach

This approach uses the third solution space representation scheme mentioned in Section 2. Each individual is represented as a completed puzzle where the third constraint mentioned in Section 1 is met: each 3x3 grid in each puzzle contains the numbers 1 to 9 exactly once. Each 3x3 grid is thus considered a gene.

A population of 100 such individuals are initialized and their finesses calculated. The best individual and fitness in the population, at each generation, are tracked. The fitness function, described in Section 2 is used, where only repetitions in the rows and columns contribute to an increase in the fitness value (by definition, there are no repeats in the grids).

The process only implements a mutation. The mutation function is implemented on each individual separately and works as follows: a 3x3 grid is randomly selected. Two unfixed cells in the grid are then randomly selected and switched. During reproduction, the mutation function is implemented on each individual in the population.

The number of mutations implemented on an individual depends on how far the CGA has progressed (how fit the fittest individual in the population is). The number of mutations implemented is equal to the fitness of the fittest individual divided by 2, rounded up. Therefore if the fittest individual is 25 then the number of mutations implemented on each individual in the population during the next generation is 13.

The belief space defined in this approach differs from the previous approach in that it applies to each 3x3 grid as opposed to each cell. The belief space consists of four categories:

- Normative belief: each 3x3 grid contains entries ranging between 1 and 9.
- Domain specific belief: each 3x3 grid can only contain entries which are integers.
- Spatial belief: each 3x3 grid must contain the integers 1 to 9 exactly once.
- Temporal belief: when a mutation is applied to a grid, it cannot alter the value of a fixed cell.

During the initialization process it is insured that the first three belief criteria are met. During mutation, it is ensured that the fourth criterion is met. In this approach, there is an additional form of influence function. As mentioned, the best fitness *influences* how many mutations take place. Here, knowledge of the solution (how close the puzzle is from being solved) has a direct impact on how the algorithm is being implemented. This is also a form of culture [9]. The algorithm terminates when the best individual has a fitness of zero i.e. the solution has been found.

## 5. Repulsive Particle Swarm Optimization

Repulsive Particle Swarm Optimization (RPSO) is a variant of classical Particle Swarm Optimization (PSO), streamlined for complex search spaces with many local minima [11]. A background to both classical PSO as well as RPSO is presented. The application of RPSO to the Sudoku problem is then discussed.

### 5.1. Particle Swarm Optimization Background

Particle Swarm Optimization is a population based optimization technique, which attempts to simulate the way birds flock towards food [11]. A group of particles carry out a stochastic search, iteratively altering each particle's velocity and position until the optimal position is reached [11]. The positions of the particles are randomly initialized. A particle's position is evaluated based on a fitness function. Like GA, the fitness function is the only part of the algorithm which has knowledge of the problem.

In classical PSO, two values are used in determining a particle's next velocity and position: the particle's personal best position and the best position achieved by any particle in the group over all iterations. A particle's velocity and position is altered as follows [11]:

$$v_{i+1} = \omega v_i + \omega c_1 r_1 (x_{best} - x_i) + \omega c_2 r_2 (g_{best} - x_i) \tag{1}$$

$$x_{i+1} = x_i + v_{i+1} \tag{2}$$

Where: $x_i$ is the current position of the particle.

$x_{i+1}$ is next position of the particle.

$v_i$ is the current velocity of the particle.

$v_{i+1}$ is the next velocity of the particle.

$\omega$ is an inertial constant, ranging between 0.01 and 0.7.

$c_1$ and $c_2$ are constants that indicate to what extent the particle moves towards the best positions.

$r_1$ and $r_2$ are random numbers ranging between 0 and 1.

$x_{best}$ is the best position found by the particle.

$g_{best}$ is the best position found by the group.

Equations (1) and (2) are iteratively applied until the algorithm is terminated. The algorithm is terminated once the global optimal point has been found or after a certain number of iterations have passed. For multi-dimensional individuals, the velocity and position of each dimension is calculated separately. This is analogous to the orthogonality of a particles position and velocity in three dimensional space.

### 5.2. Repulsive Particle Swarm Optimization Background

RPSO is a variant of PSO, which is particularly effective in finding the global optimum point in very complex search spaces [12]. This is done by causing the particles to repel one another, preventing the particles from gravitating towards local optimum points. Due to its repulsive nature, RPSO takes longer to find the optimal point in simpler search spaces, when compared to classical PSO. In RPSO, a particle's velocity is altered as follows [9]:

$$v_{i+1} = \omega v_i + \omega c_1 r_1 (x_{best} - x_i) + \omega c_2 r_2 (x_{br} - x_i) + \omega c_3 r_3 z \tag{3}$$

Where: $x_i$ is the current position of the particle.

$v_i$ is the current velocity of the particle.

$v_{i+1}$ is the next velocity of the particle.

ω is an inertial constant, ranging between 0.01 and 0.7.

$c_1$, $c_2$ and $c_3$ are constants. $c_2$ should be negative and indicates a particle's repulsion away from another randomly chosen particle's best position.

$r_1$, $r_2$ and $r_3$ are random numbers ranging between 0 and 1.

$x_{best}$ is the best position found by the particle.

$x_{br}$ is the best position found by another randomly chosen particle.

z is the velocity of another randomly chosen particle.

Equation (2) is also used to alter the particles position. The algorithm is terminated once the global optimal point has been found or after a certain number of iterations have passed.

## 5.3. RPSO Applied to the Sudoku Problem

RPSO is a suitable candidate for solving Sudoku due to its efficiency in finding the global optimum. As mentioned above, due to the constraint nature of the Sudoku problem, the avoidance of local minima is critical. RPSO uses the second solution space representation scheme mentioned in Section 2: each particle's position consists of 47 dimensions, with each dimensions corresponding to a non-fixed cell in the puzzle. Hence, as with the first CGA approach, a population of 50 individuals is chosen. The population is randomly initialized and the initial velocities are set to zero. It is ensured that all particle positions only contain integers from 1 to 9.

The personal best positions are initialized to the current values. A random best position from the population ($x_{br}$) (for the same dimension) is selected to be implemented in equation (3). The values for $C_1$, $C_2$ and $C_3$ in equation (3) are 2, -2 and 2 respectively. C2 is made negative in order to implement the repulsion property between particles. ω is set to 0.1. The first set of velocities is generated using equation (3), taking $v_i$ and z to be zero. The positions of the particles are then updated, ensuring that the positions remain integers ranging between 1 and 9. Each particle has 47 dimensions. Each dimension has a corresponding position value and velocity value. When the velocity and position of a particle is updated, each dimension's velocity and position is updated separately.

Consecutive iterations update the particle velocities and positions using equations (3) and (2), where z is a different particle's velocity (for the same dimension), randomly chosen. Even though the global best position is not used to update the velocity, it is tracked in order to check when the solution is found.

## 6. Quantum Simulated Annealing

Quantum Simulated Annealing (QSA) is a powerful optimization tool which incorporates quantum tunnelling into classical Simulated Annealing [13]. A background to both classical SA as well as QSA is presented. The application of QSA to the Sudoku problem is then discussed.

## 6.1. Simulated Annealing Background

Simulated Annealing (SA) is an optimization technique, inspired by the annealing process used to strengthen glass or crystals [14;15;16]: a crystal or glass is heated until it liquefies and then is cooled slowly, allowing for the molecules to settle into lower energy states.

Unlike GA and PSO, SA is not population based but rather alters and tracks the state of an individual, continuously evaluating its energy by using an energy function [14]. SA finds the optimal point by running a series of Markov Chains under different thermodynamic conditions [14]: a neighbouring state is determined by randomly changing the current state of the individual by implementing a neighbourhood function. If a state with a lower energy is found then the individual moves to that state. Otherwise, if the neighbouring state has a higher energy then the individual will move to that state only if an acceptance probability condition is met. If it is not met, then the individual remains at the current state.

The acceptance probability is a function of the difference in energies between the current and neighbouring states as well as the temperature [14]. The temperature is initially made high, making the individual more susceptible to moving to the higher energy state. This allows the individual to explore a greater portion of the search space, preventing it from being trapped in a local optimum. As the algorithm progresses the temperature is reduced, in accordance with a cooling schedule, causing the individual to converge towards the state with the lowest energy and hence the optimal point.

A typical SA algorithm works as follows [14]:

- Initialize an individual's state and energy.
- Initialize temperature.
- Loop until temperature is at minimum.
    - Loop until maximum number of iterations has been reached.
        - Determine neighbouring state by implementing the neighbourhood function.
        - Determine the energy of the current and neighbouring state.
        - If the neighbouring state has a lower energy than the current, then change the current state to the neighbouring state.
        - Else, if the acceptance probability is fulfilled then move to the neighbouring state.
        - Else stick to the current state.
        - Keep track of state with lowest energy.
    - End inner loop.
    - Alter temperature in accordance with the cooling schedule.
- End outer loop

There are various types of cooling schedules [13]: a linear cooling schedule simply subtracts a constant from the previous temperature. A geometric cooling schedule multiplies the previous temperature by a fraction.

### 6.2. Quantum Simulated Annealing Background

QSA is based on quantum tunnelling, where an individual can move from one state to another by tunnelling through high energy areas, in search of lower energy areas [13]. This is controlled by the tunnelling field strength (analogous to the temperature in SA). The major difference between QSA and SA is that in SA the temperature determines the probability of moving from one state to the next, while the neighbourhood remains constant. In QSA, the tunnelling field strength determines the neighbourhood radius or the distance between the current state and the neighbouring state (how different the two are) [13].

### 6.3. QSA Applied to the Sudoku Problem

QSA is used in discrete search spaces which contain many local optima [13], making it ideal for Sudoku. Since QSA is not a population based algorithm, the third solution space representation scheme mentioned in Section 2 is used. The starting state of the annealing process is initialized by randomly filling the puzzle, but ensuring that each 3x3 grid contains the integers 1 to 9 exactly once. The fitness function is described in Section 2, where only repetitions in the rows and columns contribute to an increase in the fitness value (by definition, there are no repeats in the grids).

The neighbourhood function is identical to the mutation function described in the second CGA approach: a grid is randomly selected. Two unfixed cells in the grid are then also randomly selected and switched.

The initial tunnelling strength should be set high enough to ensure that 80 % of states are accepted during the initial Markov Chain [13]. The tunnelling strength is therefore initialized by taking the standard deviation of 100 randomly initialized states and setting it to that value. 20 iterations of tunnelling strength reduction are implemented (the outer loop of the algorithm has 20 iterations). During each iteration, the tunnelling strength is reduced by multiplying it by 0.8. This ensures that the tunnelling strength decreases fast enough to ensure a solution after 20 iterations but not too fast so as to make the acceptance probability too greedy.

The Markov Chain is set to run for 2209 iterations. This is due to the fact that in general, the size of the Markov Chain should represent the size of the search space [14] which in this case is the 47 unfilled cells of the puzzle. Therefore, the Markov Chain should be $47^2 = 2209$ iterations. The probability function is defined as follows [13]:

$$P(e, e', t) = \exp\left(-\frac{e' - e}{t}\right) \quad (4)$$

Where:  e = energy of current state

   e' = energy of neighbouring state

   t = tunnelling strength

As mentioned above, QSA is different from classical SA in that the tunnelling strength also determines the neighbourhood radius. In this instance, the tunnelling strength defines how many neighbourhood operations are applied to the current state, to derive the neighbouring state. Initially, when the tunnelling strength is high, the current state will undergo many random cell swaps (within various randomly chosen grids). As the tunnelling strength decreases, less random swaps will be undertaken until only one random swap is applied during the final Markov Chain.

## 7. Hybrid Genetic Algorithm and Simulated Annealing

A hybrid of Genetic Algorithm and Simulated Annealing (HGASA) combines the parallel search capability of GA with the flexibility of SA. GA is a population based search, which allows for the search space to be explored by multiple individuals. SA allows for the possibility of finding a better optimal point, even though a local minimum has been located. Both of these aspects are essential in finding the solution for a Sudoku puzzle. A background to HGASA is presented and its application to the Sudoku problem is then discussed.

### 7.1. Hybrid Genetic Algorithm and Simulated Annealing Background

At the beginning of a search, it is necessary to explore as much of the search space as possible. A population based search is ideal for this. Hence, initially a GA is implemented until a low fitness is achieved. Once the GA cannot find any better individuals after a certain number of generations, the best individual in the population is chosen to undergo a series of random walks until the optimal solution is found.

A number of factors need to be defined when implementing a HGASA. It is necessary to determine how long the GA is required to run (how many generations). Once the number of generations is determined, a modified SA algorithm is run on the fittest individual until the solution is found.

Other aspects to consider are all the same parameters discussed above for the GA and SA. The GA population representation and size; the reproduction function and the neighbourhood function (the GAs reproduction function should be similar to or incorporate the SAs neighbourhood function) as well as the fitness function.

Due to the fact that a hybrid model is being used, it might seem fitting for certain aspects of the GA and SA algorithms, mentioned above, to be omitted. For example, the reproduction function might only implement a mutation and leave out crossover. If the GA produces an individual with a fitness close to the optimal then the temperature schedule could be omitted and only a single Markov Chain implemented.

### 7.1. HGASA Applied to the Sudoku Problem

A population of ten puzzles is initialized. The third solution representation scheme mentioned in Section 2 is used: each 3x3 grid in the puzzle contains the integers 1 to 9 exactly once. The reproduction function is the same as mentioned above for the second CGA approach: only mutation is implemented which randomly selects a grid and randomly swaps two unfixed cells in the grid. The number of mutation also corresponds to the best fitness, as per the CGA discussed in section 4.4. The best fitness and individual are tracked until a fitness of 2 is found (experimentally it was established that the GA found a fitness of 2 very quickly).

Once an individual with a fitness of 2 is found, the individual enters into the SA cycle. Since the GA managed to find an individual with a low energy, the individual can enter into the SA process fairly late; hence no temperature schedule is implemented. Therefore instead of implementing a Markov Chain, a simple moderated Monte Carlo Chain is used. The Monte Carlo Chain accepts states with energies which are lower or equal to the current state's energy and runs until a state with the energy of zero is reached (the solution is found).

The SA and GA cycles both share the same fitness function and the neighbourhood function used in the SA is the same as the mutation function used during the GA (with only one mutation applied per iteration).

## 8. Results and Technique Comparison

After testing all four algorithms on the puzzle in Figure 1, the second CGA approach, QSA and HGASA algorithms were able to find the correct solution depicted in Figure 2. The RPSO and the first CGA approach were unable to find a solution.

### 8.1. CGA Results

CGA took 28 seconds to find the solution after 208 generations (at best). Figure 3 shows a plot of the fitness of the best individual in each generation across the entire run.

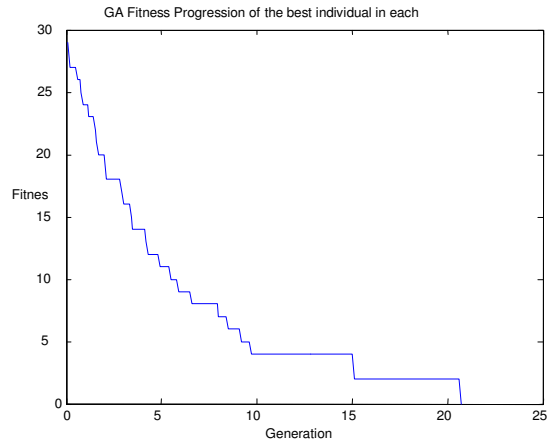

Figure 3: Fitness progression during the CGA run.

The CGA was run 20 times and was able to find the solution every time. The time taken to find the solution varied significantly from 28 seconds to 10 minutes. This is due to the fact that CGA is a stochastic process and the time to convergence depends on how close the initial population is to the solution, as well as on the random mutations applied to the individuals of the population.

As is evident from Figure 4, the GA is able to find an individual with a low fitness (less than 5) very quickly but takes significantly longer to find the optimal solution. From Figure 4, the fitness drops from 28 to 4 in less than 100 generations but it takes over 100 generations for the fitness to drop to 0. Furthermore, during each generation a population of 100 individuals has to undergo reproduction, which most likely won't result in a better individual being found. Hence, much computational effort is wasted on trying to find the solution from a low fitness, resulting in the algorithm taking longer to converge (when compared to the other algorithms). HGASA provides a solution to this problem as discussed in Section 8.3.

### 8.2. QSA Results

QSA took 65 seconds to run and found the solution after 42700 iterations. Figure 4 shows a plot of the energy reduction across the entire run.

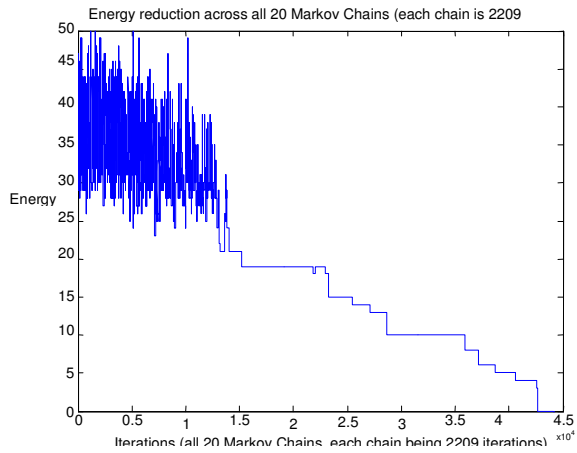

Figure 4: Energy reduction during QSA run.

The relatively long time taken to converge is primarily due to the initial iterations, when multiple random swaps are carried out. The QSA was run 20 times and found the solution only 15 times. This is due the fact that the QSA is run for a set number of iterations before stopping (hence a constant time to convergence). The result is highly dependent on the random mutations applied to the state as well as on the acceptance of higher states.

This problem could be remedied by introducing a reheat mechanism which causes the QSA to reinitialize the state and increase the temperature, or by increasing the length of the Markov Chains. Both would result in the algorithm taking longer to find the solution. As is evident from Figure 4, the initial Markov Chains accept higher energy states more often (due to the temperature schedule) and neighbouring states vary significantly from iteration to iteration. Towards the end of the run, neighbouring states vary less and fewer higher energy states are accepted, as expected. Furthermore, there is a more gradual reduction of energy as the algorithm progresses (when compared to the CGA).

Also note the tunnelling effect taking place during the later Markov Chains, where a lower energy state is searched for without moving to high energy states (the high energy states are 'tunnelled through').

### 8.3. HGASA Results

HGASA solved the puzzle (at best) in 1.447 seconds and after 435 iterations. Figure 6 shows the fitness progression across both the GA cycle as well as the SA cycle. The GA cycle achieved a fitness of 2 in 168 iterations and the SA implemented a Monte Carlo Chain of 267 iterations before finding the solution.

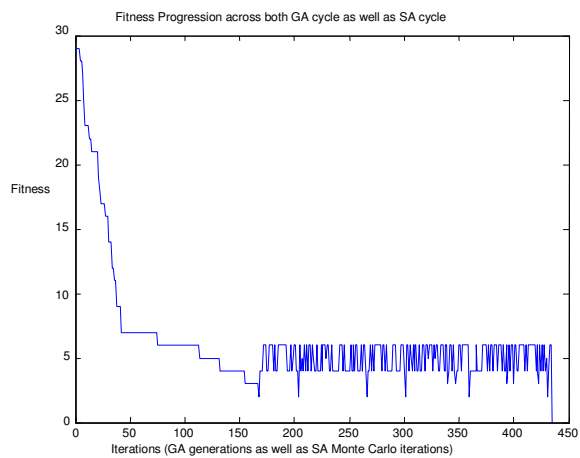

Figure 5: Fitness progression during HGASA run.

The HGASA was run 20 times and solved the puzzle every time. The time it took to find a solution took between 1.447 seconds and 3 minutes. This is due to the stochastic nature of the algorithm (like the CGA and QSA, the time to convergence depends on the initialization as well as on the random swaps).

Like with the CGA, the HGASA is able to find an individual with a very low fitness, very quickly (after 170 generation it found an individual with a fitness of 2), after which the Monte Carlo Chain takes over. The reason why the HGASA is able to find the solution faster than the CGA is because the Monte Carlo Chain is much less computationally demanding: it only operates on one individual as opposed to an entire population and hence a Monte Carlo iteration takes less time than a GA generation. The HGASA find a solution quicker than the CGA because it uses a smaller initial population and hence, even the GA cycle is computationally less demanding.

Hence, the HGASA is deemed the optimal algorithm to use to solve the Sudoku problem. The HGASA could possibly be improved by applying the Monte Carlo Chain to multiple individuals at a higher fitness.

### 8.4. RPSO and the first CGA approach

RPSO and the first CGA approach were not able to find a solution to the Sudoku problem. This is mainly due to the way the individuals and particles are represented and manipulated. RPSO is not deemed a suitable candidate for solving Sudoku since the search operations implemented cannot be naturally adapted to generating a better possible puzzle. The same applies for the search operations defined for the first CGA approach. This results in the algorithm never being able to converge to solution since better individuals are unlikely to be generated from iteration to iteration.

## 9. Conclusion

The Sudoku problem can be solved efficiently using stochastic search techniques. Four stochastic optimization algorithms were applied to a Sudoku puzzle and their implementation discussed in this paper: CGA, RPSO, QSA and HGASA. Two different CGA approaches were implemented. Solution space representation is discussed as well as the fitness function (which is common to all algorithms). A background to all the algorithms is presented; their application to the Sudoku problem discussed.

Only one of the CGA approaches, the QSA and the HGASA were able to find the solution to the puzzle. CGA found the solution in 28 seconds and after 208 generations. QSA found the solution in 65 seconds and after 42700 iterations. HGASA proved to be the most efficient algorithm, solving the puzzle in 1.447 seconds and after 435 iterations. This is because HGASA combines the parallel searching power of GA with the flexibility of SA. The HGASA can possibly be improved by implementing the SA cycle on more than one individual.